\pdfoutput=1
\documentclass{article}

\usepackage{etoolbox}
\usepackage[preprint]{neurips_2023}

\usepackage{amsmath,amsthm,amssymb,appendix,bm,graphicx,hyperref,mathrsfs}

\usepackage{fancyhdr}
\usepackage{booktabs}
\usepackage{multirow}
\usepackage{caption}
\usepackage{subcaption}
\usepackage{threeparttable}
\usepackage{todonotes}

\usepackage{setspace}
\usepackage{float}

\newtheorem{lemma}{Lemma}[section]
\newtheorem{theorem}{Theorem}[section]

\newtheorem{remark}{Remark}

\usepackage{abstract}

\usepackage[T1]{fontenc}
\usepackage[utf8]{inputenc}
\usepackage{lmodern}  

\usepackage{bbm}  

\usepackage{algorithm}
\usepackage{algpseudocode}

\title{BUZZ: Beehive-structured Sparse KV Cache with Segmented Heavy Hitters for Efficient LLM Inference}
\author{\textbf{Junqi Zhao$^{1*}$,
	Zhijin Fang$^{2*}$,
	Shu Li$^{3}$\thanks{Equal contribution.},
	Shaohui Yang$^{4}$,
	Shichao He$^{5}$ }\\
  $^1$ Department of Computer Science, New York University, New York, USA.\\
  $^2$Department of Statistics and Data Science, University of California, Los Angeles, USA. \\
  $^3$Department of Mathematical Science, Beihang University, Beijing, China.\\
  $^4$School of Computer Science, University of Leed, Leeds, United Kingdom.\\
  $^5$Department of Computer Science and Applied Math, Brandeis University, Boston, USA. \\
  \texttt{jz4840@nyu.edu, danielfang@ucla.edu, leo7090@buaa.edu.cn} \\
 }

\begin{document}
\maketitle
\begin{abstract}
Large language models (LLMs) are essential in natural language processing but often struggle with inference speed and computational efficiency, limiting real-time deployment. The key-value (KV) cache mechanism reduces computational overhead in transformer models, but challenges in maintaining contextual understanding remain. In this paper, we propose BUZZ, a novel KV caching algorithm that leverages structured contextual information to minimize cache memory usage while enhancing inference speed. BUZZ employs a beehive-structured sparse cache, incorporating a sliding window to capture recent information and dynamically segmenting historical tokens into chunks to prioritize important tokens in local neighborhoods. We evaluate BUZZ on four real-world datasets: CNN/Daily Mail, XSUM, Wikitext, and 10-QA. Our results demonstrate that BUZZ (1) reduces cache memory usage by \textbf{2.5}$\times$ in LLM inference while maintaining over 99\% accuracy in long-text summarization, and (2) surpasses state-of-the-art performance in multi-document question answering by \textbf{7.69\%} under the same memory limit, where full cache methods encounter out-of-memory issues. Additionally, BUZZ achieves significant inference speedup with a $\log{n}$ time complexity. The code is available at: \href{https://github.com/JunqiZhao888/buzz-llm}{\texttt{https://github.com/JunqiZhao888/buzz-llm}}.
\\

\end{abstract}
\section{Introduction}

In the realm of artificial intelligence, the research on Large Language Models \cite{wei2022emergent,nam2024using} has, in recent years, become one of the most compelling areas of focus. 
With the introduction of the attention mechanism \cite{vaswani2017attention}, transformer-based network structures have been proven to perform well in various fields, such as personalized content recommendation on streaming platforms, drug discovery through molecular property prediction, and enhancing customer service chatbots for more natural and accurate responses. These structures have been widely adopted in these practical scenarios.

Since that seminal work, a plethora of pre-trained models has been introduced, each leveraging extensive linguistic datasets. Prominent among these are Llama \cite{brown2020language,roziere2023code}, GPT \cite{brown2020language,achiam2023gpt}, Gemini \cite{team2023gemini}, and Mistral \cite{jiang2023mistral}, models that have demonstrated remarkable proficiency in language processing and generation.

As application scenarios continue to evolve and more models wish to tackle longer context window, they impose increasingly stringent demands on computational efficiency, storage requirements, and performance efficacy. In response to the imperative to mitigate computational overhead, the KV cache \cite{waddington2013kv} mechanism has been introduced. During the prefilling and decoding process within the forward layer in transformer structure \cite{han2021transformer,hassani2023neighborhood}, the $K$ and $V$ , denoting key and value, matrices for certain tokens are stored once computed, thus avoiding redundant calculations and substantially reducing the overall computational effort of the algorithm \cite{liu2024minicache}. Moreover, the KV cache is remarkably adaptable, perfectly aligning with various attention optimization methods such as the Sparse Attention Method \cite{wang2021spatten,lin2022swinbert}, the GPU-optimized Flash Attention Method \cite{dao2022flashattention}, and the Sketching Method within HyperAttention \cite{gribonval2021sketching,han2023hyperattention}, thereby achieving a multiplicative acceleration of processing speeds.

\begin{figure}[t]
    \centering
    \includegraphics[width=1.05\textwidth]{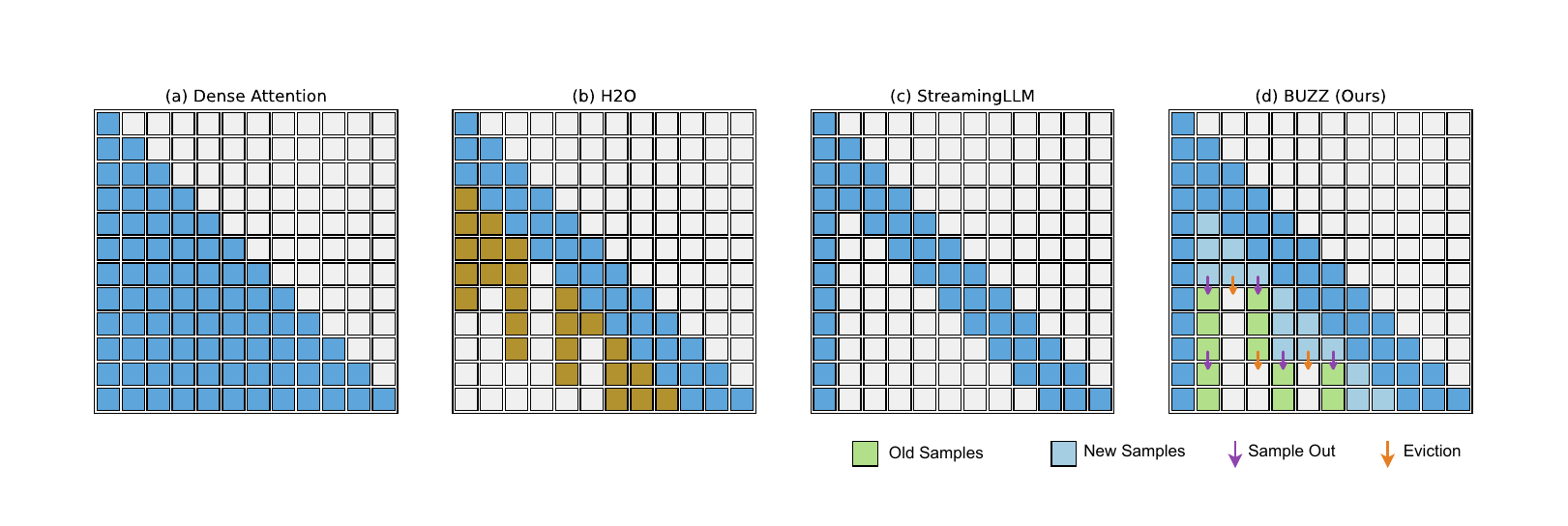}
    \caption{Illustration of BUZZ vs. existing methods: We visualize sequential decoding steps, where grey blocks represent masked or unseen tokens, and colored blocks represent retained tokens in each respective cache method. Method (a) illustrates the full cache, the default scheme that consumes substantial memory as context length increases. Method (b) dynamically retains heavy-hitters (tokens with high attention scores) outside the window. Method (c) modifies the local method by adding a narrow attention sink, which significantly enhances performance in benchmark experiments. Our method, BUZZ (d), dynamically retains heavy-hitters while preserving the contextual structure.}
    \label{f1}
\end{figure}

The efficacy of a KV cache is contingent upon its storage logic, which can significantly influence the model's performance. An effective KV cache must operate with both rapid speed and high accuracy in predicting and generating responses, while also judiciously selecting the most representative and informative elements from preceding tokens, all the while minimizing storage space utilization. However, in real-world scenarios such as information retrieval, educational tools, and decision-making aids, the utility of multi-turn and long text conversations is more prevalent (\emph{e.g.}, chatbotting, following up on retrieval results). Unfortunately, existing KV cache models, even those optimized with various sliding windows \cite{jiao2023dilateformer,duanmu2024skvq} techniques, have demonstrated sub-optimal performance in these areas.

To surmount these challenges, we introduce BUZZ, a new KV cache approach that selectively stores and evicts key-value pairs in prefilling and decoding state to maintain at a reduced capacity. Divergent from traditional Heavy Hitter \cite{zhang2024h2o} selection methodologies, our model employs a beehive-like structure, strategically selecting Heavy Hitters in local communities to achieve KV sparsity, as shown in Figure \ref{f1}.

In a comparative analysis with alternative models, the full attention caching mechanism illustrated in Figure \ref{f1}(a) incurs significant memory overhead, largely due to the retention of key-value pairs for all preceding tokens during each token generation. This approach introduces considerable computational inefficiencies and memory constraints, resulting in performance degradation and Out-Of-Memory (OOM) failures, particularly during extended dialogue interactions. As an optimization over local window strategies shown in Figure \ref{f1}(b), StreamingLLM \cite{xiao2023efficient}, depicted in Figure \ref{f1}(c), employs an attention sink and a sliding window mechanism to achieve key-value simplification. However, this comes at the cost of discarding intermediate information, which can lead to inaccuracies in responses. The H$_2$O model \cite{zhang2024h2o}, as shown in Figure \ref{f1}(d), incorporates attention scores among intermediate tokens within a sliding window framework, resulting in improved predictive accuracy to some extent. Nevertheless, this method is not without limitations: the time complexity associated with global searches increases substantially, and the omission of tokens with comparatively lower attention scores fails to adequately capture the complete distribution of attention scores within the intermediate segments \cite{noor2022simple}. Additionally, a greedy heavy-hitter selector may produce a skewed and unrepresentative attention matrix, adversely affecting model performance as content variability and context length grow.
\begin{figure}[H]
    \centering
    \includegraphics[width=0.9\textwidth]
    {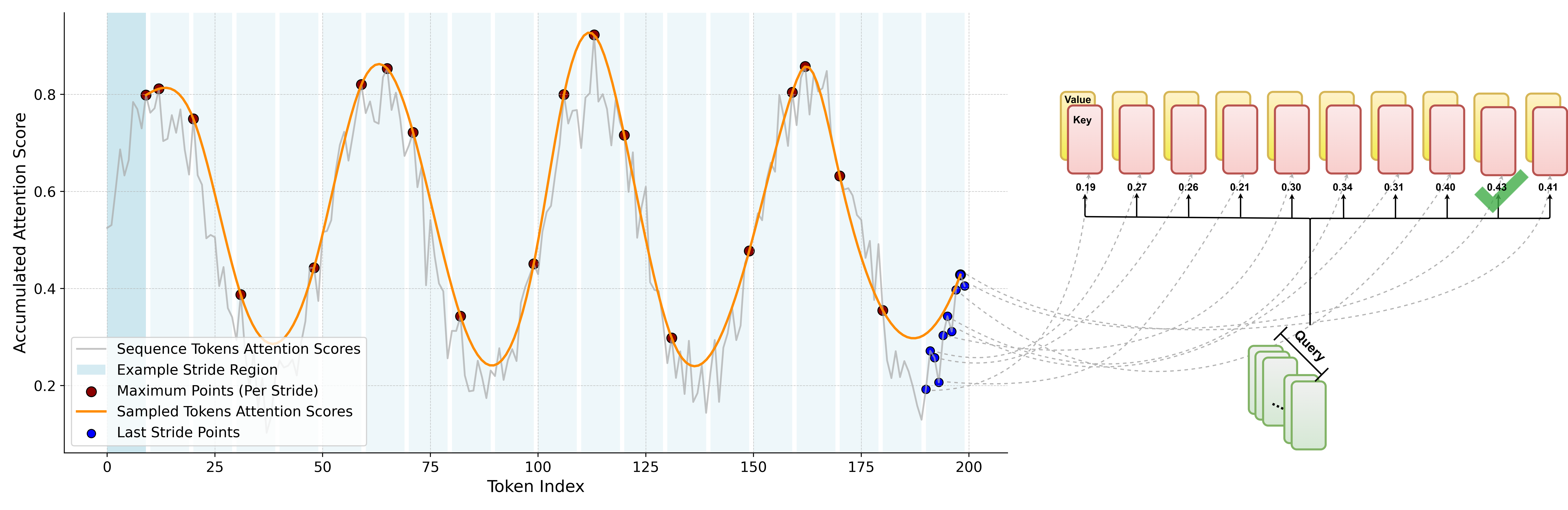}
    \caption{Overview of the local heavy hitter mechanism (BeeHive) in BUZZ: BUZZ approximates the attention scores of middle tokens, extracts local maxima, and evicts the rest. The stride size, a user-defined parameter, controls the granularity of these local neighborhoods.}
    \label{f2}
\end{figure}

BUZZ (Figure \ref{f1}(e)), building upon the aforementioned discussions, introduces a novel segmented sampling approach to identify peak values within uniform segments, offering several key advantages:\\
\textbf{Low time complexity.} BUZZ's localized search mechanism attains a time complexity of $O(n)$ and effectively exploits parallel computation, thereby significantly enhancing computational efficiency and scalability.\\
\textbf{Enhanced memory dynamics.} BUZZ modulates the step sizes between older and newer tokens, emulating human-like memory retention patterns where older tokens are represented more densely while newer tokens are sparser, thus reinforcing the retention of crucial information over time.\\
\textbf{Preservation of complete contextual information.} The $k$ tokens identified through the maximum value search within each segment extend beyond mere top-$k$ token selection. This method ensures the comprehensive retention of critical past token information while accurately restoring the distribution of attention scores by maintaining the overall shape. As shown in Figure \ref{f2}, we envision that the model must remain highly attentive to both important tokens and the overarching contextual structure to optimize performance.

\section{Related Work}
\noindent\textbf{Window methods.} In the realms of multi-turn dialogues and long-text processing, researchers have proposed various window-based strategies to address the challenges of memory consumption and computational efficiency.  There is a local attention mechanism \cite{child2019generating} that restricts the model to focus on only a portion of the sequence at any given time, thereby reducing computational load and enhancing processing speed. StreamingLLM \cite{xiao2023efficient} maintains model performance by retaining a limited number of initial token key-value pairs and integrating a sliding window mechanism. Longformer \cite{beltagy2020longformer} employs evenly spaced windowed attention mechanisms, combining local and global attention to handle lengthy texts. LM-Infinite \cite{han2023lm} effectively processes long sequences without additional learning by introducing lambda-shaped attention masks and distance constraints. MInference \cite{jiang2024minference} accelerates the prefilling phase by dynamically constructing sparse indices that recognize unique patterns in long contexts.\\
\textbf{KV Cache Eviction methods.} These methods enhance the inference efficiency of LLMs by selecting more important tokens and compressing KV cache. The H$_2$O\cite{zhang2024h2o} algorithm stands out in this field: based on the discovery that a small subset of tokens, known as "Heavy Hitters," contribute most of the value when calculating attention scores, researchers have achieved effective compression of KV cache by dynamically balancing the most recent tokens with these heavy hitter tokens. Scissorhands \cite{liu2023scissorhands} sequentially predicted the potentially pivotal tokens with the attention score above average within a history window. The Keyformer \cite{adnan2024keyformer} optimizes the scoring function of the aforementioned model, employing the Gumbel softmax method and a temperature coefficient $\tau$, which further enhances the model's performance. Additionally, the SubGen \cite{zandieh2024subgen} algorithm compresses KV cache using clustering and sampling methods while maintaining model performance. The LESS \cite{dong2024get} algorithm learns the residual between the original attention output and the sparse strategy approximation, accumulating the discarded information from the sparse strategy into a low-rank state, thus restoring the important attention areas that were neglected. \\
\textbf{KV Cache Quantization methods.} There are also several quantization methods specific to KV Cache, which are orthogonal to KV Cache Eviction methods. KVQuant \cite{hooper2024kvquant} conducts channel-based quantization for attention keys and token-based for attention values due to their distinct patterns. The low rank method employed by LESS \cite{dong2024get} for KV Cache can also be seen as one of the quantization methods. Similarly, GEAR \cite{kang2024gearefficientkvcache} applies quantization to the majority of KV Cache of similar magnitudes to ultra-low precision. It then employs a low rank matrix to approximate the quantization error, and a sparse matrix to remedy individual errors from outlier entries of KV Cache. The methods of quantization can be combined with KV Cache Eviction methods to achieve higher compression. But in our main text, we just focus on KV Cache Eviction and leave this part to future work.

\section{Methodology}
The goal of this section is to propose the algorithm of our method BUZZ and to show some preliminary and guarantees. We first present the preliminaries of KV Cache Eviction and the motivation of BUZZ. Then we detail BUZZ from an algorithmic perspective.
\subsection{Preliminary}
We first give a basic preliminary about the inference of LLMs based on transformer \cite{vaswani2017attention} layers and the problem formulation of KV Cache Eviction. Here we only consider one head attention block. Let the attention weight be $W_Q\in\mathbb{R}^{d\times d}$, $W_K\in\mathbb{R}^{d\times d}$ and $W_V\in\mathbb{R}^{d\times d}$, and the embedding of prompt or hidden state of last layer be $X\in\mathbb{R^{seq\times d}}$, where $d$ represents the hidden dimension of the model and $seq$ represents the length of prompt in prefilling part and 1 in decoding part. 

The inference of the LLMs follows the autoregressive way. KV Cache is initialied as
\begin{equation}
    K = X_{prompt}W_K, V = X_{prompt}W_V
\end{equation}
At each decoding step $t$, the model generates next token based on KV Cache and update $K$ and $V$ it into $[K,K_t]$ and $[V,V_t]$, where $K_t$ and $V_t$ are computed by
\begin{equation}
    Q_t = X_tW_Q, K_t = X_tW_K, V_t = X_tW_V 
\end{equation}
Then the attention outputs are calculated by
\begin{equation}
    A_t = Softmax(\frac{1}{\sqrt{d}}Q_tK^T), O_t = A_tV
\end{equation}
Hence, KV Cache reduces the time complexity of computation of attention into linear level based on length of sequence. However, the GPU memory usage of KV Cache is a new challenge. Here we give the formulaton of KV Cache Eviction.

At decoding step $t$, given the known matrices $Q_t, K, V$, our objective is to determine matrices $\hat{K}, \hat{V}$ that minimizes the function concerning $\hat{K},\hat{V}$ defined as the difference between two attention matrices:
\begin{equation}
    \min \left|\hat{O}(\hat{K},\hat{V}) - O_t\right|
\label{objective}
\end{equation}

where the new attention output are computed as follows:
\begin{equation}
    \hat{A} = Softmax(\frac{1}{\sqrt{d}}Q_t\hat{K}^T), \hat{O} = \hat{A}\hat{V}
\end{equation}

This minimization ensures that the modified attention output $\hat{O}(\hat{K},\hat{V})$ closely approximates the original one.

\subsection{Motivation}
Prior work\cite{xiao2023efficient} has shown that deploying a sliding window and ultilizing attention sink tokens which often refer to initial tokens to capture head and tail information can significantly reduce cache memory while maintaining model performance. Additionally, preserving top-$k$ tokens based on attention scores $A$ outside the sliding window which are referred to as heavy hitters has been found to enhance model potential without increasing cache memory usage\cite{zhang2024h2o}. However, these methods do not fully leverage the coherence and structure of natural language, as they apply a standard greedy algorithm to select important tokens, often overlooking the holistic importance of past tokens in terms of structure. 

Moreover, previous studies\cite{adnan2024keyformer} have explored evicting KV Cache and reconstructing an ideal attention score distribution, assuming it should follow a Gumbel distribution, which is a skewed left Gaussian distribution that favors earlier tokens and penalizes later ones. Although this approach introduces a bias that tends to preserve structured information, BUZZ offers a more flexible method by segmenting tokens outside the sink tokens and sliding window to identify local heavy-hitters, thus compressing KV Cache while protecting structured attention distribution and recovering attention distribution without distorting its inherent shape\cite{dong2024get}. 
This segmentation ensures that token generation attends to every part of the preceding text, avoiding excessive focus on localized tokens. Furthermore, prior KV Cache eviction methods usually evict one token immediately after decoding a new one, making the accumulated cost of step-by-step eviction non-negligible. Considering this, at each eviction step, BUZZ removes tokens from each segment and introduces a threshold parameter, allowing the model to buffer KV Cache of tokens and conduct eviction only when the threshold is reached, reducing the additional overhead of the eviction operation and enhancing the efficiency of algorithm.

Our approach enables the model to utilize the information from most of the tokens from different parts including head, tail, and structured body while still limiting cache memory. And according to the objective in \ref{objective}, our aim is to achieve a significant increase in computational speed while minimizing the loss of information. This is accomplished by retaining the structured and informational content of $K,V$ in optimized matrices $\hat K,\hat V$ as much as possible via our method, BUZZ.

\subsection{BUZZ Algorithm}
\begin{figure}[H]
    \centering
    \includegraphics[width=1.0\textwidth]{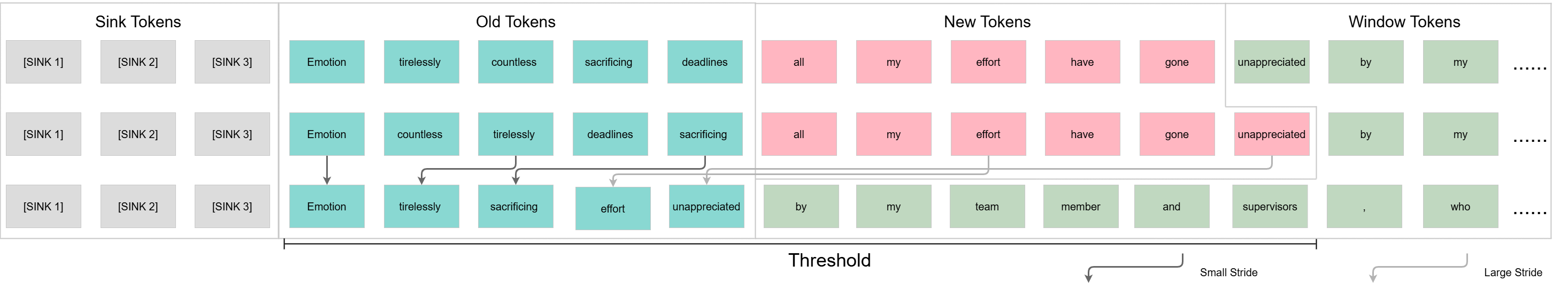}
    \caption{Algorithm Illustration: New tokens are placed in the buffer. Once the total token count reaches the threshold, we sample the new tokens with a large stride and the old tokens with a small stride. After this process, all current tokens become the old tokens, and the buffer is cleared to accommodate new tokens.}
    \label{picture}
\end{figure}

Our approach consists of three components that address the head, body, and tail of the cached KV, each employing distinct strategies for pruning before concatenating them to form a updated cache. Before describing our algorithm, we first list the relavant notations in Table \ref{notation} which we will explain in detail. As Figure \ref{picture} illustrates, the KV Cache of BUZZ has three main features:

\begin{table}[H]
\centering
\begin{tabular}{|c|c|}
\hline
\textbf{Symbol}        & \textbf{Description}                                   \\ \hline
$s$            & Big sample stride for new tokens       \\
$\hat{s}$ & Small sample stride for old tokens                  \\
$k$            & Sink tokens size                                 \\
$w$            & Sliding window size                                   \\
$T$            & Threshold indicating one eviction  \\
$A$      & Attention scores for tokens               \\
$N$            & Input context length                          \\ \hline
\end{tabular}
\vspace{5pt}
\caption{A comprehensive list of symbols and abbreviations employed throughout the text.}
\label{notation}
\end{table}
\vspace{-5pt}

\noindent\textbf{Attention sink.} Empirical evidence\cite{xiao2023efficient} supports the use of narrow attention sink tokens, which outperforms standalone window methods. We adopt this approach, set the size of the sink tokens in $k$, which is often a small number ranging from 1 to 5 and find that it stabilizes our method's performance, confirming its effectiveness throughout our study.

\textbf{Sliding window.} The sliding window approach is a standard technique, capturing information from recent tokens to ensure that relevant context is retained during generation. This method allows the model to focus on the most recent context, maintaining coherence and relevance in the generated text. For long-sequence tasks, such as summarization or question-answering, the sliding window acts as a minimum guarantee, working in tandem with our BeeHive sampling mechanism. We carefully tuned the window size $w$ to balance sufficient context retention with minimized computational overhead.

\textbf{BeeHive.} To counteract the greedy selection of tokens with high attention scores, we introduced the BeeHive module, which segments intermediate tokens based on the predefined stride $s$ into hives to limit the scope of selection. In each hive, we conduct the eviction and only retain one token based on the attention scores, which we can regard as a single sampling. We separately evaluate 'new' tokens, those recently exited the sliding window, and 'old' tokens that have persisted through multiple evictions. Recognizing the importance of old tokens, after the first eviction, we reduce the stride by half to $\hat{s}=\lfloor \frac{s+1}{2}\rfloor$ in the subsequent eviction steps to capture more historical information. This segmentation ensures the recovery of the overall attention distribution after eviction due to the integration of structured information. 

Here we detail a description of how the KV Cache Buffering and Eviction work in BUZZ:

\textbf{Buffering Strategy.} As we mentioned earlier, BUZZ does not carry out eviction step by step. Each eviction reduces the size of the KV cache outside the sink tokens and sliding window by approximately $\frac{s-1}{s}$. We implement a threshold $T$ that buffers the incoming tokens. These tokens are first buffered at the end of the sliding window and stored as part of the new tokens once the sliding window has moved away. When the threshold $T$ is reached, new tokens are segmented using the user-defined stride $s$, while old tokens are segmented with half the stride size $\hat{s}$. These segments are then aggregated into the old tokens for the next decoding step.

Next, we present our detailed eviction policy algorithmically in algorithm \ref{kv_cache_algorithm} and the local max sampling method in algorithm \ref{localmax}. Local max samling method is designed for new tokens while for old tokens we directly utilize interval sampling to preserve structured information better, which just select items every $\hat{s}$ intervals. The relevant notations in this chapter are shown in Table \ref{notation} and we have explained some of them in the preceding context.
\begin{remark}
	In the described policy, we assume that the number of tokens exceeds 
	$k+w$. If the number of tokens is insufficient, priority is given to filling the sink and window first.
\end{remark}
\begin{remark}
 In fact, the eviction policy is specifically tailored for the decoding phase. In the prefilling phase, if the input length is too long, we will continuously conduct sampling until the selected tokens can fit within a container of length $T+k+w$.
\end{remark}

\begin{algorithm}
\caption{KV Cache eviction algorithm of BUZZ}
\label{kv_cache_algorithm}
\hspace*{0.02in} {\bf Input:} Sink token size $k$, Sliding window size $w$, Stride $s$, Current Cache $K, V$, Old tokens after last eviction $K_{old},V_{old}$, New sequence $(K_1, V_1),(K_2, V_2),\cdots,(K_t, V_t)$ , Threshold $T$ for sampling\\
\hspace*{0.02in} {\bf Output:} Updated Cache $\hat{K}, \hat{V}$ after getting new-coming tokens and conducting one eviction, old tokens for next eviction $K_{old},V_{old}$
\begin{algorithmic}[1]
\State $n\gets size(K)$
\State $i\gets1$
\While{$n \leq k+w$}
\State $K\gets[K, K_i], V\gets[V,V_i]$
\State $n\gets n+1, i\gets i+1$
\EndWhile
\State $n_{old}\gets size(K_{old})$
\State $m\gets0$
\While{$m \leq T$}
\State $K\gets[K, K_i], V\gets[V,V_i]$
\State $m\gets m+1, i\gets i+1$
\EndWhile
\State $\hat{s}\gets \lfloor \frac{s+1}{2}\rfloor$
\State $K_{new}\gets K[k+n_{old}:-w], V_{new}\gets V[k+n_{old}:-w]$
\State $K_{old},V_{old}\gets \mathbf{interval\_sample}(\hat{s},K_{old},V_{old})+\mathbf{local\_max\_sample}(s,K_{new},V_{new})$
\If{$i < t$}
\State {goto $\mathbf{7}$ and $\mathbf{repeat}$}
\EndIf
\State $\hat{K}\gets \big[K[:k],K_{old},K[-w:]\big],\hat{V}\gets \big[V[:k],V_{old},V[-w:]\big]$
\end{algorithmic}
\hspace*{0.02in} {\bf Return:} $\hat{K},\hat{V},K_{old},V_{old}$
\end{algorithm}

\begin{algorithm}
\caption{Local max sample algorithm}
\label{localmax}
\hspace*{0.02in} {\bf Input:} Stride s, Cache $K,V$, Accumulated attention scores $A$ corresponding to $K,V$\\
\hspace*{0.02in} {\bf Output:} Updated cache $K_{res}, V_{res}$ after sampling
\begin{algorithmic}[1]    
\State Set $K_{res}, V_{res}$ empty
\State $n\gets size(K), idx\gets 0$
\While{$idx<n$}
\State $m\gets argmax(A[idx:idx+s])$
\State $K_{res} = \big[K_{res},K[m]\big], V_{res} = \big[V_{res},V[m]\big]$
\State $idx\gets idx+s$
\EndWhile
\end{algorithmic}
\hspace*{0.02in} {\bf Return:} $K_{res},V_{res}$
\end{algorithm}
\textbf{Tima Complexity.} Since the attention score matrix is derived from the output of the model's original decoding layer, the time complexity of computing the attention scores is excluded from the analysis of our KV update algorithm. Our approach performs $O(1)$ operations for chunking and identifying each local heavy hitter, resulting in an overall time complexity of $O(n)$.

\subsection{Parameter Estimation}
In our algorithm, there are numerous parameters, including $k, w, s, T$, and selecting the appropriate ones is a crucial issue. The following theorem provides some guidance on the selection of these parameters before conducting eviction.
\begin{theorem}[Parameter Estimation Theorem]
	Maintaining a constant stride $s$ and cache size $C$, the performance of the LLM is expected to be optimal when the following condition is satisfied($T$ denotes the threshold for eviction, $w$ denotes sliding window size):
	\begin{equation}
		\frac{T}{w} =
		\begin{cases}
			\displaystyle\frac{s^2 + 1}{s + 1}, & \text{if } s \text{ is odd}; \\
			s-1, & \text{if } s \text{ is even}.
		\end{cases}
	\end{equation}
\end{theorem}

\begin{figure}[H]
	\centering
	\includegraphics[width=0.5\textwidth]{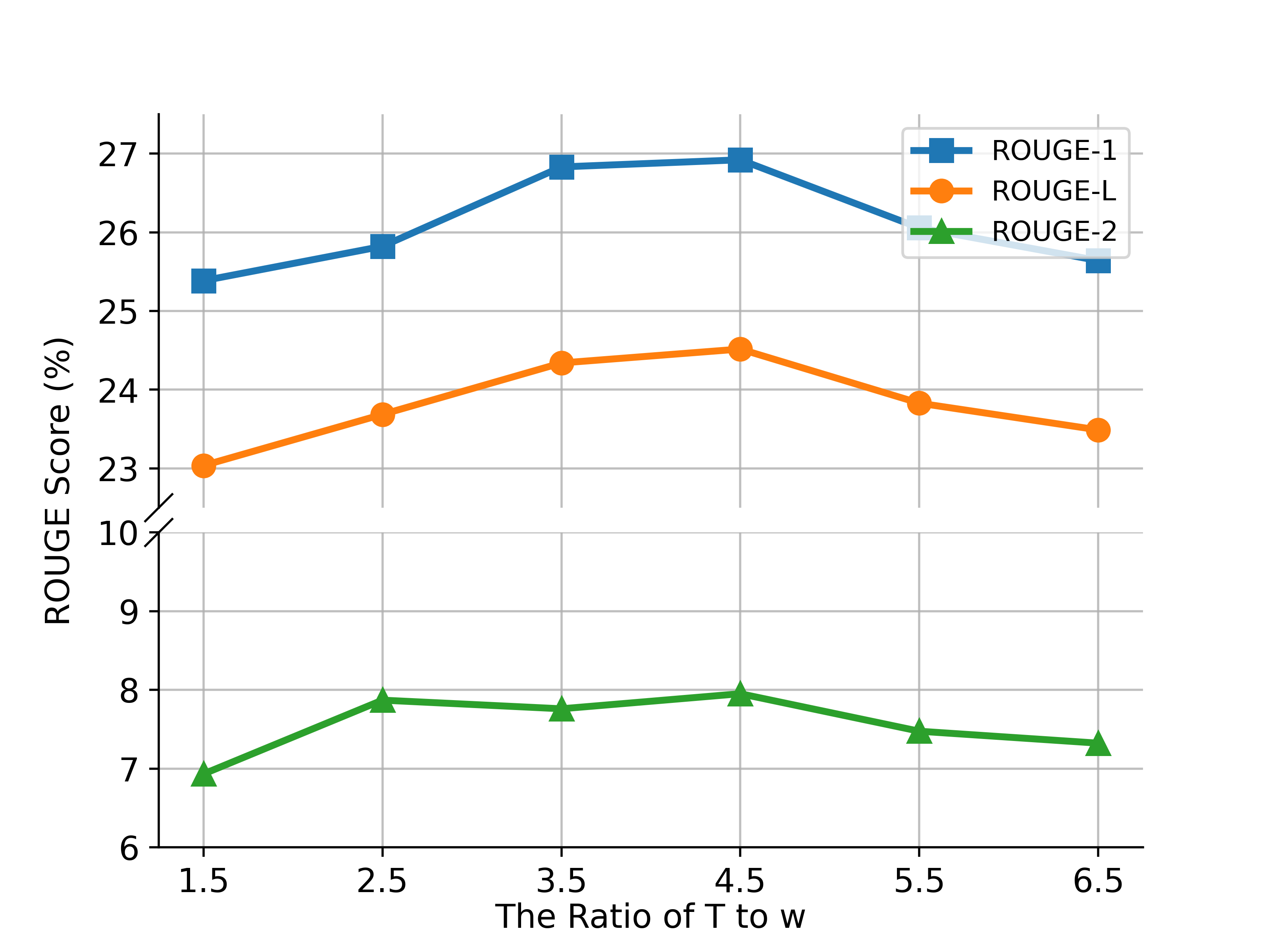}
	\caption{Model performance under different $\frac{T}{w}$ values. We choose CNNDaily as our datasets, set stride to be 5 and cache size to be around 200.}
	\label{ratio}
\end{figure}
The proof of this theorem is presented in Appendix \ref{proof of parameter estimation}. It offers guidance for the selection of parameters in BUZZ. Figure \ref{ratio} demonstrates the variation in ROUGE scores of our method under different values of $\frac{T}{w}$, given a fixed stride $s$ and cache size $C$. The detailed data are demonstrated in Appendix \ref{Tw}. Via applying our theorem, we deduced that the optimal predictive performance occurs at $\frac{T}{w} = 4.33$. The experimental outcomes corroborate our prediction, with the best performance observed at $\frac{T}{w} = 4.5$. Consequently, the parameter estimation theorem holds substantial practical significance for the application of our method.

\subsection{Augmented Inference with log n}
Many researchers believe that the softmax function in the attention mechanism of all Transformer models has certain drawbacks, such as forcing each attention head to annotate, even when there is no information to be added to the output vector\cite{xiao2023efficient}. To address this issue, we have optimized the function with $\log{n}$\cite{chiang2022overcoming} to obtain the BUZZ with logn model. To demonstrate the importance and efficiency of this technique, we first state a proposition: \textbf{In order to enable the model's results to generalize better to unknown lengths, the design of the Attention mechanism should maintain entropy invariance as much as possible.}
We then prove the following lemma:
\begin{lemma}
	Let \( t_1, t_2, \ldots, t_n \) be a set of data points, and define the probability \( p_i \) associated with each data point \( t_i \) as
	\[ p_i = \frac{e^{\lambda t_i}}{\sum\limits_{j=1}^n e^{\lambda t_j}} \]
	where \( \lambda > 0 \) is a hyperparameter. Then the entropy \( H(p) \) of the probability distribution \( p = (p_1, p_2, \ldots, p_n) \) is minimized when \( \lambda \) is proportional to \( \log{n} \).
\end{lemma}

The proof of this lemma is presented in Appendix \ref{proof of logn}. In the lemma, $\lambda$ is actually the coefficient in front of $QK^T$, and $p_i$ represents the components of softmax($QK^T$). Therefore, we have sufficient reason to add a logn coefficient. As for the determination of the base number, we believe that when the length is the mainstream model's pre-training length ($n=512$), our optimization formula should degrade to the traditional attention formula. Therefore, we take 512 as the base number, and the specific augmented attention formula is:
\begin{align}
		A(Q,K,V) &= Softmax\left(\frac{\log_{512} n}{\sqrt{d}}QK^T\right) V
\end{align}
Then we combined it with the KV Cache updated policy and get the following formula:
\begin{align}
		A(Q,\hat{K},\hat{V}) &= Softmax\left(\frac{\log_{512} n}{\sqrt{d}}Q\hat{K}^T\right) \hat{V}
\end{align}
BUZZ with $log{n}$ will serve as an auxiliary to our main BUZZ method and augment the performance of inference in various scenarios.
\section{Evaluation}
The goal of this section is to show the competitive performance of our method BUZZ. We first describe the evaluation settings and datasets. Then we present the testing results in various scenarios in detail. We also do some ablation studies to show the robustness of BUZZ.
\subsection{Settings}
\textbf{Setup.} Our experiments are based on the famous and representative model family of LLMs, LLaMA2\cite{touvron2023llama} with different model sizes. By default, we all use LLaMA2-7B. We use a total of 4 datasets which focus on long-context tasks, including CNN Daily \cite{see-etal-2017-get}, XSUM \cite{zhang2019pegasus}, Wikitext \cite{merity2016pointer}, and 10-document-QA \cite{liu2024lost}. We followed official HuggingFace/GitHub guidelines for data preprocessing and loading.

\textbf{Baselines.} The baselines we compare BUZZ with include H$_2$O which use greedy accumulated attention scores to evict KV Cache of non-important tokens, StreamingLLM which use recomputation of position embeddings based on sink tokens and sliding window, and local sliding window method. The upper bound of these methods is dense attention with full KV Cache.

\textbf{Evaluation Metrics}
To comprehensively assess BUZZ's performance on long-text tasks, we evaluated it across different metrics, such as ROUGE-1, ROUGE-2, ROUGE-L, perplexity, Self-BLEU, and EM accuracy. For each task, we conducted extensive hyperparameter tuning and applied theoretical approximations to post-eviction cache size, thereby reducing the number of hyperparameters.

\textbf{Implementations Details.} BUZZ is implemented on major open-source transformer frameworks by modifying their inference forward function. Our experiments were selectively conducted on NVIDIA A100 (40GB GPU RAM) and NVIDIA L4 GPUs (22.5GB GPU RAM), depending on the required query context length of each task.
\subsection{Experiments}
\subsubsection{Summarization}
\begin{figure}[H]
    \centering
    \includegraphics[width=1\textwidth]{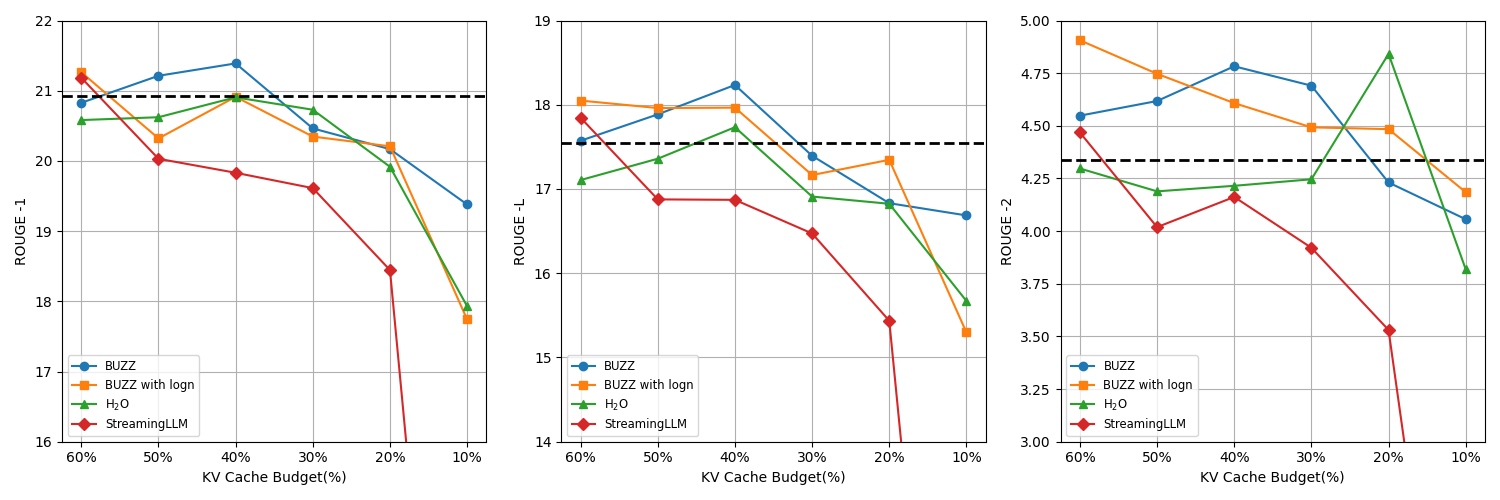}
    \caption{Comparison of methods on summarization ROUGE scores vs KV cache budget. The black dotted line embodies the accuracy achieved by full cache method and is marked by the dotted lines in the graphs.}
    \label{f4}
\end{figure}

\textbf{Task description.} Summarization tasks are crucial for evaluating a language model's ability to distill long articles into concise, coherent summaries while preserving essential information. To assess BUZZ's effectiveness in handling long-context scenarios, we employed a widely recognized dataset: XSUM\footnote{In the experiments conducted on the XSUM dataset, the highest ROUGE scores of the local window model is significantly lower than others in all three charts, and therefore is not displayed in the figure.}, offering unique challenges and serving as a robust framework for this evaluation.

\textbf{Results.} Figure \ref{f4} illustrates the comparison of ROUGE scores versus cache budget among BUZZ, H$_2$O, StreamingLLM, and local method. Cache Budget is defined as the percentage of the average cache size used by the full model. ROUGE scores for each method were recorded at predetermined cache thresholds to evaluate the consistency and effectiveness of cache memory utilization. Mean ROUGE is calculated as the geometric mean of ROUGE-1, ROUGE-2, and ROUGE-L scores. 

Figure \ref{f4} demonstrates that BUZZ and BUZZ with $\log{n}$ consistently achieve the highest ROUGE scores, and the detailed data is shown in Appendix \ref{XSUM}. Notably, BUZZ reaches and even exceeds the 99\% accuracy baseline with only about 40\% of the original cache size. Moreover, the smooth, gradually decreasing trend across all three metrics indicates that BUZZ is the least susceptible to performance degradation under constrained cache budgets. A detailed analysis of the geometric mean ROUGE versus cache budget is provided in Table \ref{mean_score}, where BUZZ exhibits superior summarization capabilities, effectively utilizing precise diction, generating coherent bi-grams, and restructuring content with concise language. We also validate that BUZZ outperforms other alternative methods on Llama2-13B, and the detailed results are available in Appendix \ref{13b}.

\begin{table}[H]
\caption{Mean Score Comparison between different models. The bold numbers are the highest mean rouge score under the given KV cache  budget. The result shows that our BUZZ model outperforms the H$_2$O, StreamingLLM model overall.}
\label{mean_score}
\centering 
\begin{tabular}{ccccccc}
\toprule 
Mean Score(\%)  & 60\%   & 50\%   & 40\%   & 30\%   & 20\%   & 10\%   \\
\midrule 
BUZZ & 11.9 & \textbf{12.1} & \textbf{12.3} & \textbf{11.9} & 11.3 & \textbf{10.9} \\
BUZZ (with logn) & \textbf{12.3} & 12.0 & 12.0 & 11.6 & 11.6 & 10.4 \\
H$_2$O & 11.5 & 11.4 & 11.6 & 11.4 & \textbf{11.8} & 10.2 \\
StreamingLLM & 11.9 & 11.1 & 11.2 & 10.8 & 10.0 & 3.8 \\
Local & 7.6 & 4.4 & 2.9 & 1.8 & 1.4 & 0.1 \\
\bottomrule 
\end{tabular}
\end{table}

\subsubsection{Context-based Q\&A}
\begin{figure}[htbp]
    \centering
    \includegraphics[width=0.5\textwidth]{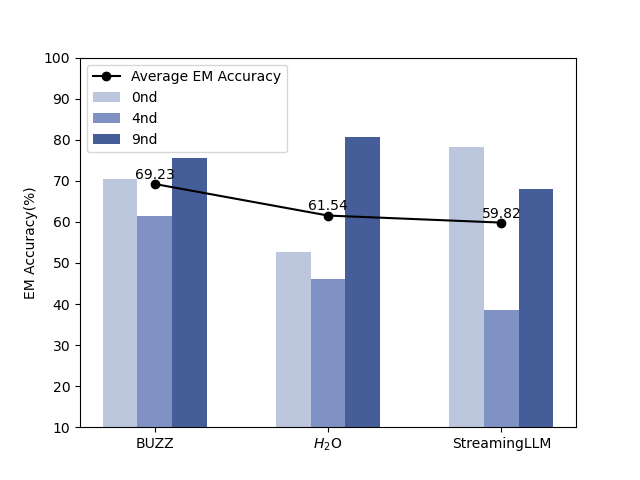}
    \caption{Comparison of BUZZ, H$_2$O and StreamingLLM on multi-document question-answering exact matches accuracy. Without requiring any task-specific fine-tuning, our model stays consistently at the high level of accuracy regardless of the position of the answer in the context. H$_2$O and StreamingLLM suffered significant loss of information in the middle.}
    \label{f5}
\end{figure}
\textbf{Task description.} Context-based QA is a common scenario in online streaming, where context may refer to user-provided documents or multiple chunks of conversation history. Therefore, a model's ability to extract relevant information from any part of the context is critical. In our evaluation, we used the multi-documents QA dataset \cite{liu2024lost}, where each task includes 10 contextual documents of similar size and a question, with the correct answer located in a controlled position within one of the documents. 

\textbf{Testing details.} We set the cache size threshold to 300, approximately 25\% of the average input token count. The full cache method encountered OOM issues on NVIDIA L4 on the first question and was therefore excluded from this comparison. 

\textbf{Results.} Figure \ref{f5} compares the performance of different methods on three types of multi-document QA datasets, where the reference answers are located at position 0, 4, and 9. We also computed the average EM accuracy to assess overall performance on long-sequence question answering. BUZZ significantly outperformed other methods in terms of average EM accuracy, particularly in position 4 QA datasets, and achieved relatively high scores in both position 0 and position 9 datasets. This superior performance indicates BUZZ's consistency and applicability in real-life QA scenarios, where target answers may be located at the beginning, middle, or end of the provided context.

\textbf{Analysis.} As mentioned above, BUZZ outperforms other methods, especially in position 4 QA datasets. This result further confirms the importance of structured information, which BUZZ utilizes, while H$_2$O employs a special mechanism to capture important tokens outside its window module, the high-variance loss suggests that the window plays a larger role in retaining information than keeping top-k heavy hitters in practical usage. StreamingLLM, which focuses mechanically on the head and tail by discarding middle information, performs as expected according to its approach.

Also, BUZZ's low-variance loss ensures that as the number of documents or the length of the context increases, the model's ability to capture key information scales effectively. Additionally, we anticipate that traditional heavy-hitter methods will see a significant drop in accuracy when dealing with 20 or more documents. As the reference answer becomes sparser in the context, this method could become increasingly unpredictable, often capturing only narrow, localized information.
\begin{figure}[H]
    \centering
    \includegraphics[width=0.6\textwidth]{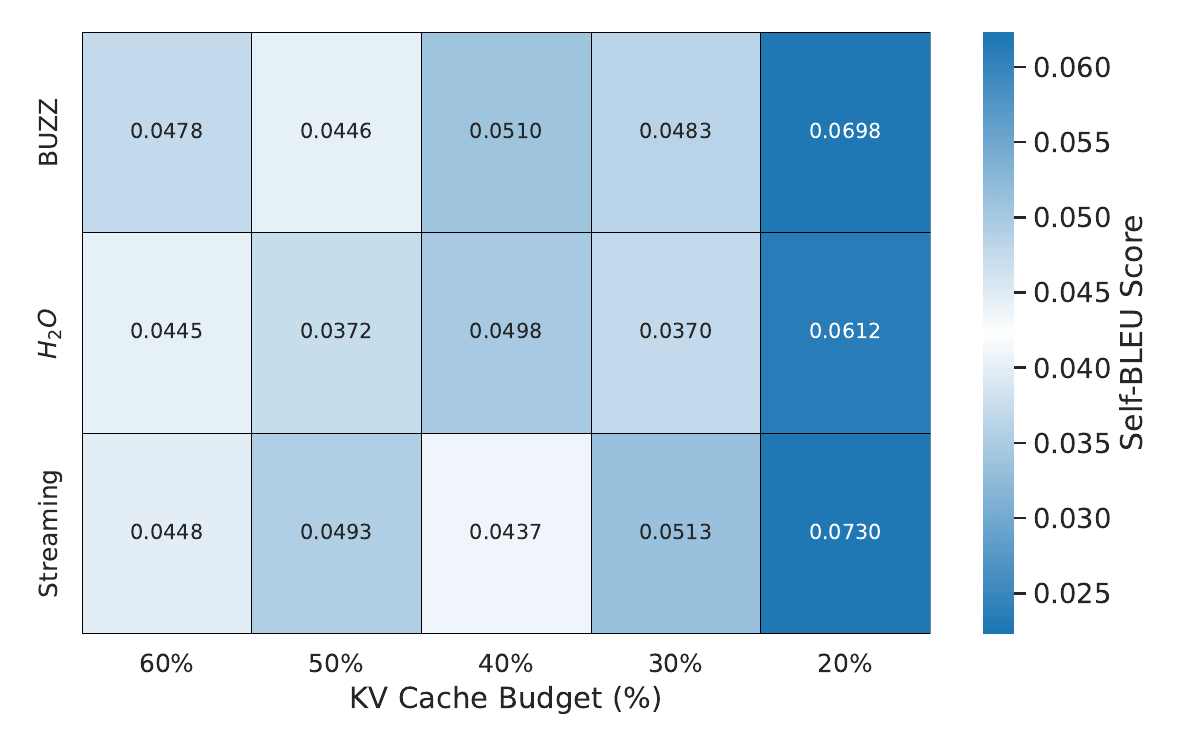}
    \caption{Heatmap based on Self-BLEU score in BUZZ, BUZZ with logn, H$_2$O and StreamingLLM. The shallower the color, the closer is the value to Self-BLEU score of full attention. BUZZ demonstrates consistent and controlled diversity despite reducing cache budget. We also see a significant reduce in language diversity when cache budget shrinks to 20\%.}
    \label{fig:example}
\end{figure}
\subsubsection{Diversity and Perplexity}

\textbf{Diversity.} The Self-BLEU metric quantifies the degree of similarity among sentences generated by a model, commonly utilized to assess the diversity in language generation tasks, particularly for LLMs. 

This metric is derived by computing the BLEU scores between pairs of generated sentences. The diversity of a sentence is reflected by the average BLEU score obtained when considering other sentences as reference texts. Herein, we believe that the closer a model's Self-BLEU score is to that of a full attention mechanism, the better the predictive performance of the model \cite{holtzman2019curious} because self-BLEU score is an intrinsic property of the text itself. The greater the divergence between the model-generated text and full attention, the more likely it is that there is a significant deviation from the semantic content of the text. If a model's Self-BLEU score significantly deviates that of full attention, it indicates that its diversity is not fully grounded in textual content, and consequently, the predictive outcome is likely to be limited.
\textbf{Results.} Anchored on this hypothesis, we selected a 10\% to 60\% KV cache budget and obtained the Self-BLEU scores for BUZZ and other methods, as illustrated in Figure \ref{fig:example}. Generally, our model exhibits a performance akin to H$_2$O in terms of Self-BLEU; in contrast to StreamingLLM, BUZZ demonstrates lower scores when the KV cache budget is below 30\%, suggesting a reduced susceptibility to the decrement of cache size. However, it is worth noting that although using $\log{n}$ prominently enhanced our method’s performance on long context tasks, BUZZ with $\log{n}$ has less diversified language compared with vanilla BUZZ.

\textbf{Perplexity.} Perplexity quantifies how well a language model predicts a sample. It is defined as the exponentiation of the entropy of the model’s probability distribution over the next word, serving as a direct measure of uncertainty. Lower perplexity indicates better predictive performance, allowing researchers to compare different models or configurations effectively. Also, a model with low perplexity demonstrates a stronger understanding of linguistic structure and context. 

\textbf{Results.} Hence, to assess the uncertainty of predicting a sequence and the language understanding capability of the model ultilizing our method BUZZ, we measured the perplexity of various methods under identical cache sizes, as delineated in Table \ref{perplexity}. The results indicate that BUZZ exhibits optimal perplexity in both scenarios, with its values being closest to the level of full attention.
\begin{table}[H]
\caption{Perplexity Results for Different Models}
\centering
\begin{tabular}{ccccccc} 
\toprule 
Cache Size & BUZZ & BUZZ (with logn) & H$_2$O & StreamingLLM & Local & Full \\
\midrule 
50 & 9.394 & 12.322 & 9.987 & 11.722 & 194.803 & \multicolumn{1}{c}{\multirow{2}{*}{7.382}} \\
100 & 8.037 & 10.318 & 8.881 & 9.330 & 35.979 &  \\ 
\bottomrule 
\label{perplexity}
\end{tabular}
\end{table}

\subsection{Ablation Study}
\textbf{Importance of BUZZ BeeHive sampling.} The sampling module of BUZZ is of paramount importance. When compared against the geometric mean ROUGE score, StreamingLLM without the sampling module exhibited an average decrease of 17\% relative to BUZZ under various KV Cache budgets. Conversely, the H$_2$O model, which employs a top-$k$ module based on accumulated attention scores instead of the sampling module, demonstrated an average reduction of 4\%.

In order to further investigate the mechanisms of BUZZ, we conduct an analysis of the mechanisms underlying the stride and the local max function.

\begin{figure}[H]
    \centering
    \begin{subfigure}[b]{0.5\textwidth}
        \includegraphics[width=\textwidth]{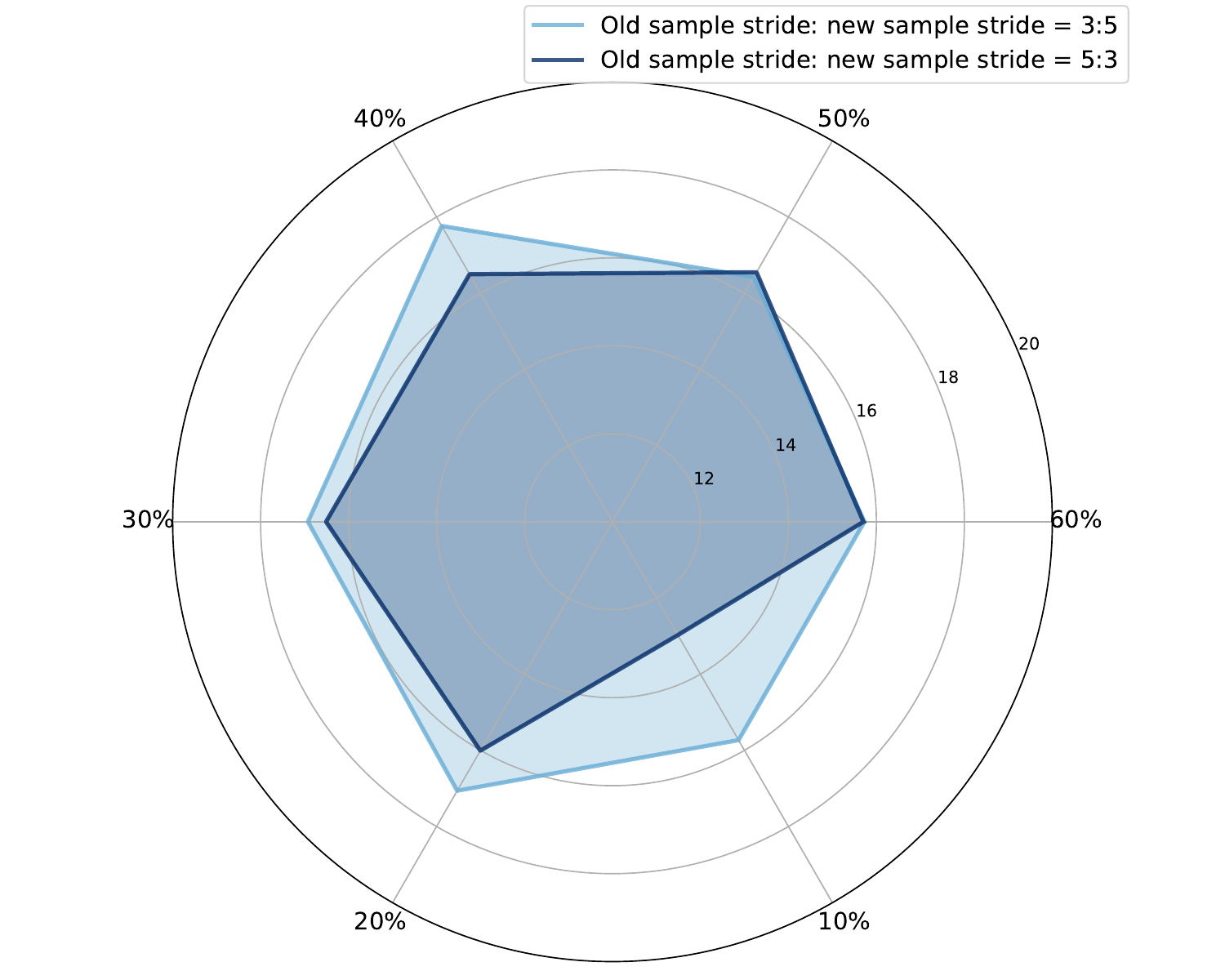}
        \captionsetup{justification=raggedright}
        \caption{}
        \label{subfig:stride_or_not}
    \end{subfigure}\hfill
    \begin{subfigure}[b]{0.48\textwidth}
        \includegraphics[width=\textwidth]{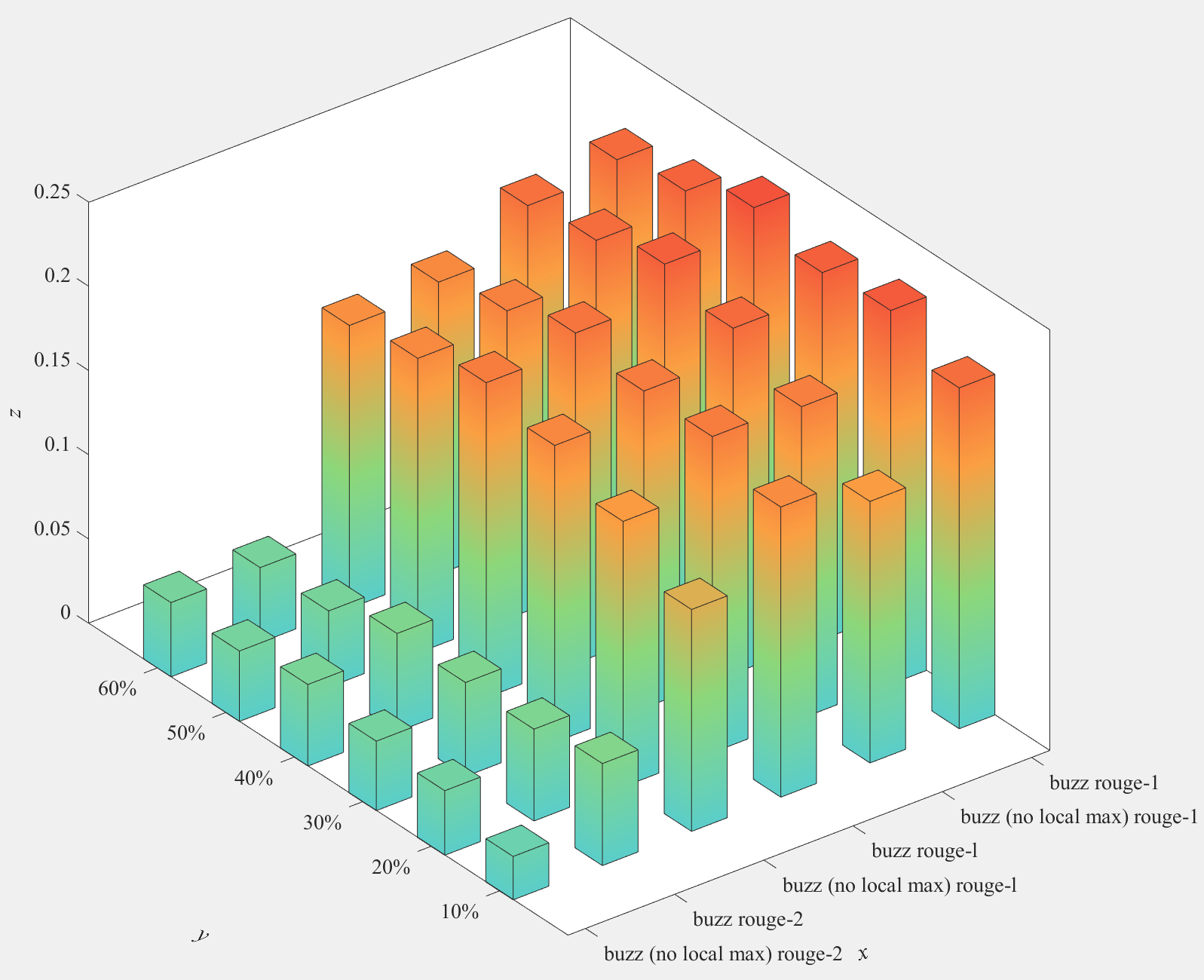}
        \captionsetup{justification=raggedright}
        \caption{}
        \label{subfig:no_max}
    \end{subfigure}
    \caption{(a) Comparison based on swapped strides: the composite ROUGE scores depicted in the figure are calculated using formula. (b) Comparison based on local-max-sample function.}
    \label{fig:image_comparison}
\end{figure}

\textbf{Influence of small stride for old tokens.} We swapped the value of small stride $\hat{s}$ and pre-defined stride $s$, measured the corresponding composite ROUGE scores for the new parameter set across 10\%-60\% of the KV Cache budget, and compared them with the old parameters. The results are depicted in Figure \ref{fig:image_comparison}(a). It is obvious that as the KV Cache budget decreases, the performance of BUZZ with swapped strides drops sharply. Theoretically, the main purpose of introducing the separate sampling size is to recover the original distribution of the attention scores. Giving new tokens a smaller stride would cause the system to favor retaining new tokens and quickly discarding old ones, thereby disrupting the distribution. 
\textbf{Influence of local max sample function.} In the current BUZZ, we employ local max sampling for new tokens and interval sampling for old tokens. To detect the necessity, we used both interval sampling for old and new tokens, and presented the results in the form of a three-dimensional bar chart, as shown in Figure \ref{fig:image_comparison}(b). Overall, as the KV Cache budget decreases, the ROUGE scores for the BUZZ (no local max) dropped rapidly. Regardless of the size of the KV Cache, the ROUGE scores for the BUZZ (no local max) were significantly lower than those of the original BUZZ. This indicates that selecting tokens corresponding to local peaks in attention scores is necessary.

\section{Conclusion}
The KV Cache reduction problem in long context processing and multi-turn dialogues has long been a focal point of research in LLMs. Our study reveals the limitations of window methods in accessing information from extensive contexts and the inefficacy of previous sparse method in accurately capturing overall contextual structure. In response, this paper introduces BUZZ, a KV cache method that reconstructs sequence attention distributions and utilizes a locally greedy selection of important tokens, all maintained within $\log{n}$ time complexity. By replacing global search with local search, BUZZ significantly reduces complexity while preserving the original distribution of attention scores as much as possible. Extensive evaluations demonstrate BUZZ's superior performance across various tasks. For example, in long-article summarization, BUZZ achieves a 2.5$\times$ reduction in cache size while exceeding over 99\% accuracy. In multi-document question answering, BUZZ outperforms state-of-the-art methods by 7.69\%. These results substantiate the practicality of our novel approach to sub-optimally selecting important tokens for KV caching.\\
\section{Acknowledgement}
We would like to thank Professor David Woodruff for sparking new ideas of inference optimization and Professor Beidi Chen for connecting us with first authors. We also thank Zhenyu Zhang for helping with reproducing H$_2$O experiment results. 

\bibliography{ref}
\newpage
\appendix
\section*{\centering Appendix}
\setcounter{table}{0}
\renewcommand{\thetable}{\Alph{table}}
\section{Text-generation Comparison of Specific Case }
\begin{figure}[h]
\centering
\vspace{-2mm}
\includegraphics[width=\linewidth]{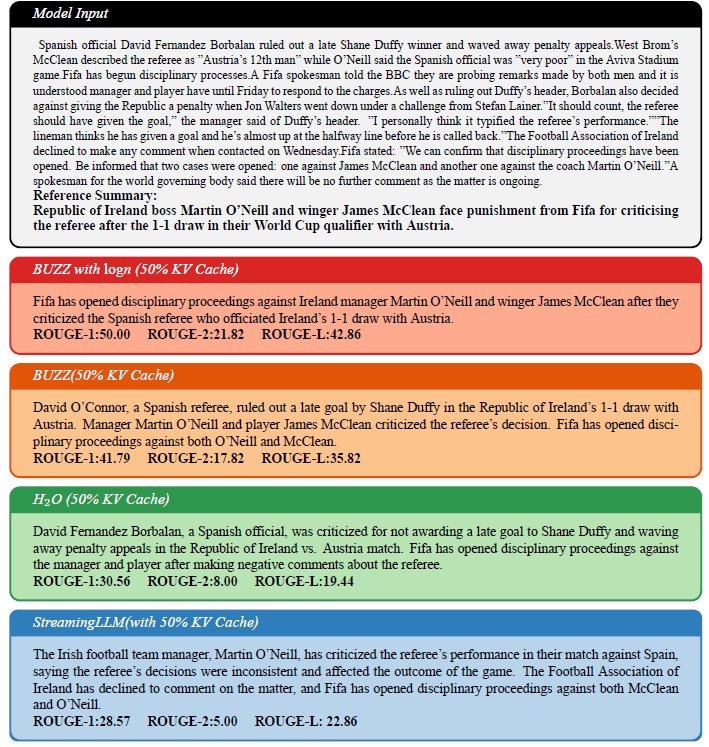}
\caption{The presented examples were run on the XSUM dataset with a 50\% KV cache budget. It can be observed that the summarization capability of StreamingLLM is significantly weaker than that of other models, as it tends to extract too many information points. The H$_2$O model performs slightly better but does not emphasize the information point that the match ended in a draw. In contrast, our BUZZ and BUZZ (with logn) models cover all the information points in shorter length, demonstrating superior summarization ability.}
\vspace{-2mm}
\label{fig:example_opt}
\end{figure}

\section{Extended Related Work, Limitation and Future Work}
\subsection{Extended Related Work}
\noindent\textbf{Quantization methods.} Quantized KV cache can minimize the impact on generation quality while reducing the memory usage of LLM in long text generation scenarios, thereby providing customizable trade-offs between memory efficiency and generation speed. QAQ \cite{dong2024qaq} combines outlier handling and attention-aware techniques, applying different quantization strategies to key and value caches, achieving a 10x compression ratio. PTQ \cite{shang2023post} focuses on accelerating the generation process by compressing noise estimation networks, maintaining performance after quantization. The WKVQuant framework adopts past-only quantization and 2D quantization strategies for quantizing weights and KV caches to save memory. KVQuant \cite{hooper2024kvquant} introduces per-channel pre-RoPE key quantization, non-uniform quantization, and dense-sparse quantization, addressing the challenge of accurate quantization at less than 4-bit precision. GEAR achieves near-lossless high compression ratios by combining quantization, low-rank matrix approximation, and sparse matrix techniques.\\
\textbf{Low-rank approximation.}  Recent research has primarily focused on enhancing the computational efficiency and adaptability of LLMs . A novel subset selection algorithm \cite{woodruff2023new} has been employed for low-rank approximations in both online and offline models. By refining the sensitivity sampling algorithm, an online core-set algorithm for $L_p$ subspace approximation has been realized. For the Schatten-p low-rank approximation problem \cite{kacham2024faster}, optimizing the rectangular matrix multiplication algorithm can significantly reduce the runtime of the algorithm. The HyperAttention mechanism \cite{han2023hyperattention}, leveraging techniques such as local sensitivity hashing, employs the computation of an approximate attention matrix to decrease the time complexity, effectively addressing the issue of approximate attention calculation in long contexts.
\subsection{Limitation and Future Work}
\noindent\textbf{Latency enigma.} While formal mathematical proofs indicate that BUZZ employs a $\log{n}$ time complexity eviction policy, comparable to other state-of-the-art algorithms, we have not yet empirically demonstrated $\log{n}$ real-time latency due to limited computational resources. In future work, we aim to parallelize computations within the BUZZ algorithm and optimize its implementation to achieve $\log{n}$ latency in practice.\\
\textbf{Future directions.} Because BUZZ is highly compatible with other inference optimization modules such as quantization methods and IO-aware optimization algorithms, we could carry out in-depth performance analysis of BUZZ in conjunction with other inference acceleration methods to achieve further performance improvements.
\section{Detailed Experiment Results}
\begin{table}[htbp]
\caption{Detailed Performance Comparision on different ratio of T to w based on constant stride 5 and cache size 200}
\setstretch{0.9}
\label{Tw}
\centering 
\begin{tabular}{ccccccc}
\toprule 
$\frac{T}{w}$  & 1.5   & 2.5   & 3.5   & 4.5   & 5.5  & 6.5  \\
\midrule 
ROUGE-1(\%) & 25.3847 & 25.8219 & 26.8306 & 26.9212 & 26.0562 & 25.6403 \\
ROUGE-2(\%) & 6.9315 & 7.8697 & 7.7600 & 7.9514 & 7.4780 & 7.3231 \\
ROUGE-L(\%) & 23.0359 & 23.6857 & 24.3394 & 24.5161 & 23.8286 & 23.4886 \\
\bottomrule 
\label{new_score}
\end{tabular}
\end{table}

\begin{table}[H]
	\caption{Detailed Summarization Performance on Llama-2-7B Model} 
 \setstretch{0.9}
   \label{XSUM}
	\centering                 
	\begin{tabular}{ccccccc}                         
		\toprule         
		\multirow{2}{*}{Method}&                     
		\multicolumn{3}{c}{\textbf{BUZZ}}&           
		\multicolumn{3}{c}
  {\textbf{BUZZ with $\mathbf{\log}$ n}}\cr     
		\cmidrule{2-4}\cmidrule{5-7}               
		&ROUGE-1 &ROUGE-2 &ROUGE-L&ROUGE-1 &ROUGE-2 &ROUGE-L\\
		\midrule         
	60\%&20.828&4.548&17.573&21.262&4.906&18.050\\
	50\%&21.212&4.618&17.886&20.321&4.747&17.961\\
	40\%&21.389&4.783&18.234&20.913&4.608&17.966\\
	30\%&20.462&4.691&17.394&20.347&4.493&17.165\\
 20\%&20.168&4.231&16.831&20.208&4.484&17.346\\
 10\%&19.379&4.056&16.686&17.742&4.185&15.298\\
		\bottomrule      
	\end{tabular}
 	\begin{tabular}{ccccccc}                        
		\toprule        
		\multirow{2}{*}{Method}&                      
		\multicolumn{3}{c}{\textbf{H$\mathbf{_2}$O}}&            
		\multicolumn{3}{c}
  {\textbf{StreamingLLM}}\cr     
		\cmidrule{2-4}\cmidrule{5-7}               
		&ROUGE-1 &ROUGE-2 &ROUGE-L&ROUGE-1 &ROUGE-2 &ROUGE-L\\
		\midrule                              
	60\%&20.582&4.297&17.106&21.185&4.468&17.846\\
	50\%&20.622&4.189&17.359&20.030&4.019&16.877\\
	40\%&20.906&4.215&17.733&19.831&4.162&16.871\\
	30\%&20.729&4.246&16.910&19.612&3.921&16.472\\
        20\%&19.916&4.841&16.824&18.446&3.531&15.436\\
        10\%&17.930&3.819&15.671&6.796&0.923&6.110\\
		\bottomrule                         
	\end{tabular}
\end{table}

\begin{table}[H]
	\caption{Detailed Summarization Performance on Llama-2-13B Model}   
 \setstretch{0.9}
   \label{13b}
	\centering                 
	\begin{tabular}{ccccccc}                         
		\toprule         
		\multirow{2}{*}{Method}&                     
		\multicolumn{3}{c}{\textbf{BUZZ}}&           
		\multicolumn{3}{c}
  {\textbf{BUZZ with $\mathbf{\log}$ n}}\cr     
		\cmidrule{2-4}\cmidrule{5-7}               
		&ROUGE-1 &ROUGE-2 &ROUGE-L&ROUGE-1 &ROUGE-2 &ROUGE-L\\
		\midrule         
		40\%&18.584&4.104&15.900&19.322&4.075&16.110\\
		30\%&17.537&4.159&15.274&19.776&4.765&16.767\\
		20\%&16.831&3.588&14.002&19.387&4.693&16.774\\
		10\%&11.941&2.149&10.083&19.159&4.558&16.233\\
		\bottomrule      
	\end{tabular}
 	\begin{tabular}{ccccccc}                        
		\toprule        
		\multirow{2}{*}{Method}&                      
		\multicolumn{3}{c}{\textbf{H$\mathbf{_2}$O}}&            
		\multicolumn{3}{c}
  {\textbf{StreamingLLM}}\cr     
		\cmidrule{2-4}\cmidrule{5-7}               
		&ROUGE-1 &ROUGE-2 &ROUGE-L&ROUGE-1 &ROUGE-2 &ROUGE-L\\
		\midrule                              
		40\%&18.677&4.287&15.780&16.666&3.168&13.675\\
		30\%&19.628&4.952&17.008&15.678&3.024&13.199\\
		20\%&17.898&4.111&14.699&15.319&2.106&12.737\\
		10\%&11.647&2.328&9.921&11.062&1.318&9.675\\
		\bottomrule                         
	\end{tabular}
\end{table}

\section{Proof of Theorems and Lemmas}
\subsection{Proof of Parameter Estimation Theorem}
	\textbf{Theorem}. Maintaining a constant stride $s$ and cache size $C$, the performance of the LLM is expected to be optimal when the following condition is satisfied($T$ denotes the threshold for eviction, $w$ denotes sliding window size):
	\begin{equation}
		\frac{T}{w} =
		\begin{cases}
			\displaystyle\frac{s^2 + 1}{s + 1}, & \text{if } s \text{ is odd}; \\
			s-1, & \text{if } s \text{ is even}.
		\end{cases}
	\end{equation}
\begin{proof}[\textbf{Proof}]
\label{proof of parameter estimation}
   According to our eviction policy , we can easily define the sequence $\{a_n\}$ recursively by our eviction rules:
	\begin{align}
		a_n = \left\lfloor \frac{2a_{n-1}}{\lfloor \frac{s+1}{2}\rfloor} \right\rfloor + \left\lfloor \frac{T-a_{n-1}}{s} \right\rfloor, \quad a_1 = \left\lfloor \frac{T}{s} \right\rfloor.
	\end{align}
	Assume that $s$ is an even integer.Using the property that for any real number $x$, $x - 1 < \left\lfloor x \right\rfloor \leq x$, we establish the following inequalities for $a_n$:
	\begin{align}
		\frac{2a_{n-1}}{s+1}  +  \frac{T-a_{n-1}}{s}  - 2 < a_n \leq  	\frac{2a_{n-1}}{s+1}  +  \frac{T-a_{n-1}}{s}.
		\label{eq:floor}
	\end{align}
	Note that the sequence $\{a_n\}$ is bounded. Taking the superior and inferior limits in (\ref{eq:floor}), we obtain:
	\begin{align}
		 \frac{T}{s}  - 2 &< \frac{s^2+1}{s^2+s}\varlimsup\limits_{n \to \infty} a_n\leq  \frac{T}{s}.
		\label{eq:infsup}
	\end{align}
	in the top-k model of H$_2$O, for long-input conversations, an empirical result is that the model performs best when $k$ is of the same order of magnitude as the window size. Therefore, we deduce that our BUZZ model performs best when $\varlimsup\limits_{n \to \infty} a_n=w$, that is:
	\begin{align}
	\frac{s^2+1}{s+1}\leq	\frac{T}{w} < \frac{s^2+1}{s+1}+\frac{2s}{w}.
	\end{align}
	If $s$ is an odd integer, by employing a similar technique, we can deduce that:
	\begin{align}
		s-1\leq	\frac{T}{w} < s-1+\frac{2s}{w}.
	\end{align}
    Since stride $s$ is generally much smaller than the window size $w$ in practice, we can neglect $\displaystyle\frac{2s}{w}$ and take the value on the left side of the inequality as the prediction value for $\displaystyle\frac{T}{w}$, thereby reaching the conclusion stated in the problem.
\end{proof}
\subsection{Proof of log n Lemma}
\label{proof of logn}
	\textbf{Lemma}. Let \( t_1, t_2, \ldots, t_n \) be a set of data points, and define the probability \( p_i \) associated with each data point \( t_i \) as
	\[ p_i = \frac{e^{\lambda t_i}}{\sum\limits_{j=1}^n e^{\lambda t_j}} \]
	where \( \lambda > 0 \) is a hyperparameter. Then the entropy \( H(p) \) of the probability distribution \( p = (p_1, p_2, \ldots, p_n) \) is minimized when \( \lambda \) is proportional to \( \log{n} \).
\begin{proof}[\textbf{Proof}]
	The entropy \( H(p) \) of the probability distribution \( p = (p_1, p_2, \ldots, p_n) \) is given by
	\begin{align}
		H(p) = -\sum_{i=1}^n p_i \log p_i = \log n + \log \left( \frac{1}{n} \sum_{i=1}^n e^{\lambda t_i} \right) - \lambda \sum_{i=1}^n p_i t_i.
	\end{align}
	For the second term, we use the approximation of taking the logarithm of the average followed by the exponential:
	\begin{align}
		\log \left( \frac{1}{n} \sum_{i=1}^{n} e^{\lambda t_{i}} \right) \approx \log \exp \left( \frac{1}{n} \sum_{i=1}^{n} \lambda t_{i} \right) = \lambda \bar{t},
	\end{align}
	where \( \bar{t} \) is the average of \( t_i \)'s.\\
	For the third term, since the softmax computation emphasizes the largest value in the data, we have
	\begin{align}
		\lambda \sum_{i=1}^{n} p_{i} t_{i} \approx \lambda \max_{i} (p_i t_i).
	\end{align}
	Thus, the entropy can be approximated as
	\[ H(p) \approx \log n - \lambda (\bar{t} - \max_{i} (p_i t_i)). \]
	To achieve entropy stability, which aims to minimize the impact of the length \( n \), we require that the derivative of the entropy with respect to \( n \) is close to zero:
	\[ \frac{\partial H}{\partial n} = \frac{1}{n} - \frac{\partial \lambda}{\partial n} (\bar{t} - \max_{i} (p_i t_i)) \approx 0. \]
 	Hence, we conclude that \( \lambda \) is proportional to \( \log n \).
\end{proof}
\end{document}